\documentclass{article}
\usepackage{spconf,amsmath,graphicx}
\usepackage{amssymb}
\usepackage{bbm}
\usepackage{dsfont}
\usepackage{bbold}
\usepackage{mathbbol}
\usepackage{newtxmath}
\usepackage{multirow}
\usepackage{lipsum}
\usepackage{float}
\usepackage{stfloats}
\usepackage{subcaption}
\usepackage{caption}
\usepackage[hidelinks]{hyperref}
\usepackage{url}
\usepackage{xcolor}
\usepackage{array}
\usepackage{booktabs}
\title{SparseSpikformer: A Co-Design Framework for Token and Weight Pruning in Spiking Transformer}
\name{Yue Liu \qquad Shanlin Xiao$^{\star}$ \qquad Bo Li \qquad Zhiyi Yu $^{\star}$\thanks{$^{\star}$ The Corresponding authors.}}
\address{Sun Yat-sen University, China}
\begin{document}
%\ninept
%
\maketitle
\begin{abstract}
As the third-generation neural network, the Spiking Neural Network (SNN) has the advantages of low power consumption and high energy efficiency, making it suitable for implementation on edge devices. More recently, the most advanced SNN, Spikformer, combines the self-attention module from Transformer with SNN to achieve remarkable performance. However, it adopts larger channel dimensions in MLP layers, leading to an increased number of redundant model parameters. To effectively decrease the computational complexity and weight parameters of the model, we explore the Lottery Ticket Hypothesis (LTH) and discover a very sparse ($\ge$90\%) subnetwork that achieves comparable performance to the original network. Furthermore, we also design a lightweight token selector module, which can remove unimportant background information from images based on the average spike firing rate of neurons, selecting only essential foreground image tokens to participate in attention calculation. Based on that, we present SparseSpikformer, a co-design framework aimed at achieving sparsity in Spikformer through token and weight pruning techniques. Experimental results demonstrate that our framework can significantly reduce 90\% model parameters and cut down Giga Floating-Point Operations (GFLOPs) by 20\% while maintaining the accuracy of the original model.
% \footnote{The code is available at \textcolor{blue}{\url{https://github.com/Rocherster/SparseSpikformer}}.}

\end{abstract}
\section{Introduction}
\label{sec:intro}

Spiking Neural Network (SNN) represents a novel type of neural network characterized by low power consumption and sparsity \cite{maass1997networks}. Different from traditional Artificial Neural Network (ANN), SNN simulates the spike transmission process of biological neurons, using discrete spike sequences for information exchange. Due to the event-driven and sparse characteristics of SNN, it can significantly reduce computing and storing resources, making it ideal for edge devices to implement \cite{akopyan2015truenorth,davies2018loihi,pei2019towards}. In recent years, researchers have been exploring optimization techniques for deep SNN \cite{hu2021spiking,zheng2021going,fang2021incorporating,deng2022temporal} to achieve better performance and efficiency, which drives the application and development of SNN in various fields.

Recently, as researchers have continued to explore Transformer architecture \cite{vaswani2017attention}, it has been gradually applied to a wide range of computer vision tasks \cite{dosovitskiy2020image,touvron2021training,carion2020end,zheng2021rethinking,wang2022detr3d}. The Vision Transformer (ViT) \cite{dosovitskiy2020image} is the first to apply Transformer to image classification tasks by transforming the input image into a sequence of patches. Since then, the ViT model and its variants \cite{touvron2021training,jiang2021all,liu2022swin} have achieved a serious state-of-the-art (SOTA) performance. The remarkable representation ability of Transformer has inspired researchers to combine it with SNN. Zhou first applied the Spiking Self-Attention in Transformer to train Spikformer \cite{zhou2022spikformer}, which broke the bottleneck of traditional SNN limitations and achieved significant success. However, its complex network architecture with high computational and storage costs has hindered its deployment on resource-constrained edge devices. To address this issue, we explore Spikformer sparsification techniques that optimize model performance and reduce power consumption by minimizing non-essential computations and weight parameters.

In this paper, we introduce SparseSpikformer, a hardware-friendly Spikformer model that adopts a co-design framework of image token and weight parameter pruning. Through this close collaboration, SparseSpikformer can achieve a substantial reduction in both computational and storage costs while maintaining a comparable level of accuracy. In summary, our main contributions are listed as follows:

\begin{itemize}
    \item We study a novel hybrid pruning framework for Spikformer with image token and weight level optimizations, to efficiently reduce hardware resource requirements and accelerate inference runtime.
    
    \item At the image token level, we adopt the average spike firing rate as a criterion and employ a token selector module to reduce the number of unnecessary image tokens. With this module, only image tokens with a high spike firing rate are kept for attention calculation, which can effectively cut down the computational cost.
    
    \item At the weight parameter level, we utilize the sparsity property of SNN to perform weight parameter pruning by LTH technique \cite{frankle2018lottery}. The experimental results demonstrate the existence of winning tickets within SparseSpikformer, encompassing only 10\% of the model parameters while exhibiting a minimal decrease in model accuracy. 
\end{itemize}

\begin{figure*}[htb]
\ninept
\centering
\centerline{\includegraphics[scale=0.565]{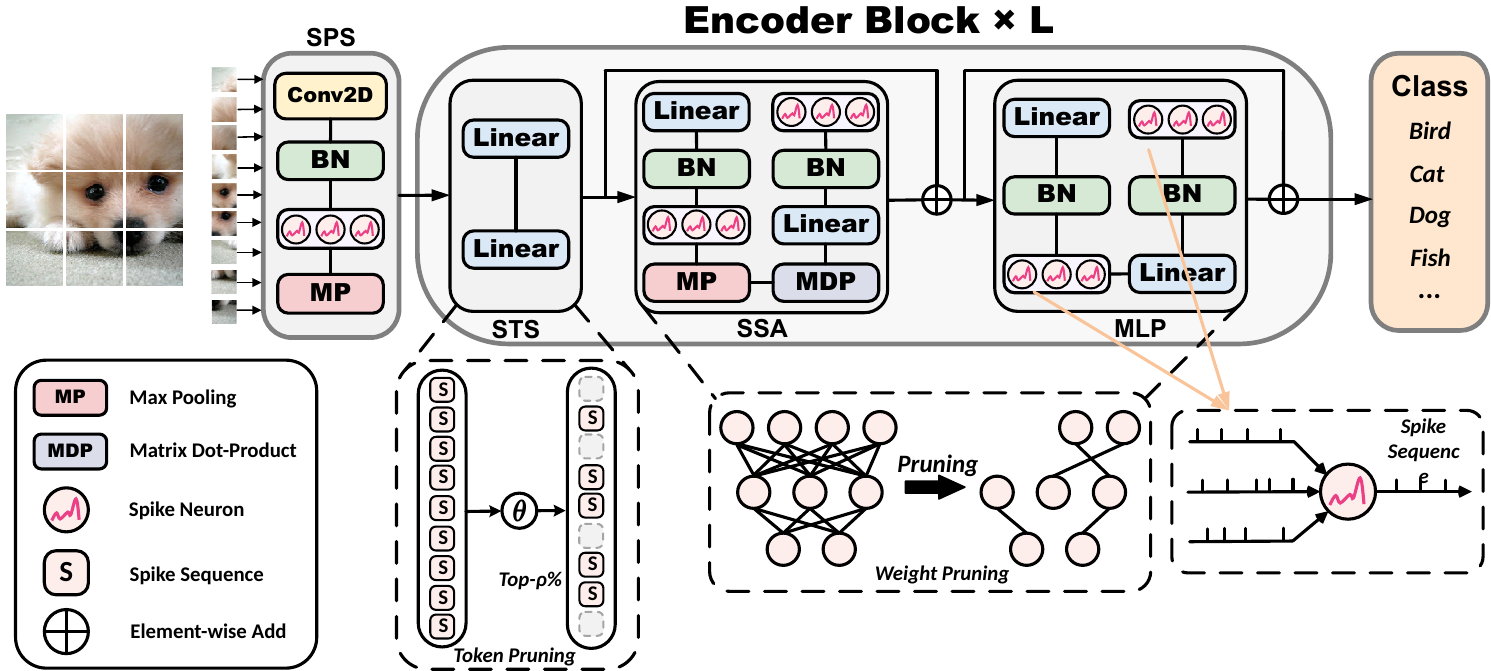}} 
\captionsetup{labelsep=period}
\vspace{-0.1cm}
\caption{The framework of SparseSpikformer. The Spiking Patch Splitting (SPS) partitions the input image into a series of spike tokens. Subsequently, the Spiking Token Selector (STS) selects the top $\rho\%$ of tokens based on their spike firing rate to participate in Spiking Self-Attention (SSA) and MLP calculations.}
\end{figure*}

\vspace{-0.4cm}
\section{Method}
\label{sec:method}

For a given 2D input image sequence $I\in \mathbb{R}^{T \times C \times H \times W}$, we follow the processing method of Spikformer. Initially, we employ Spiking Patch Splitting (SPS) to partition the input image into a series of spiking patches $X'\in \mathbb{R}^{T \times N \times D}$. Subsequently, the input spike proceeds through L layers of the Encoder Block, which consists of three essential components: Spiking Token Selector (STS), Spiking Self-Attention (SSA), and MLP. Within the STS module, we select a fixed number of tokens based on the firing rate of the spike sequence. The SSA module serves as the central component of Spikformer, primarily utilized to extract comprehensive global information from the image. It combines the characteristics of SNN by converting the traditional floating-point $Q, K, V$ matrix into spike form to reduce energy consumption. After the SSA module, the spike will further go through the MLP layer. It is worth noting that as the dimension of MLP increases, the final model accuracy will be further improved. However, this operation will also bring a large number of weight parameters, consequently increasing the computational burden. Therefore, the MLP stands out as one of the primary targets for weight pruning. After passing through L layers of the Encoder Block, we use a fully connected layer as a linear classification head to output the result $Y$. The Overall architecture of SparseSpikformer is shown in Figure 1.

\begin{figure}[tb]
\ninept
  \centering
  \includegraphics[width=1.0\linewidth]{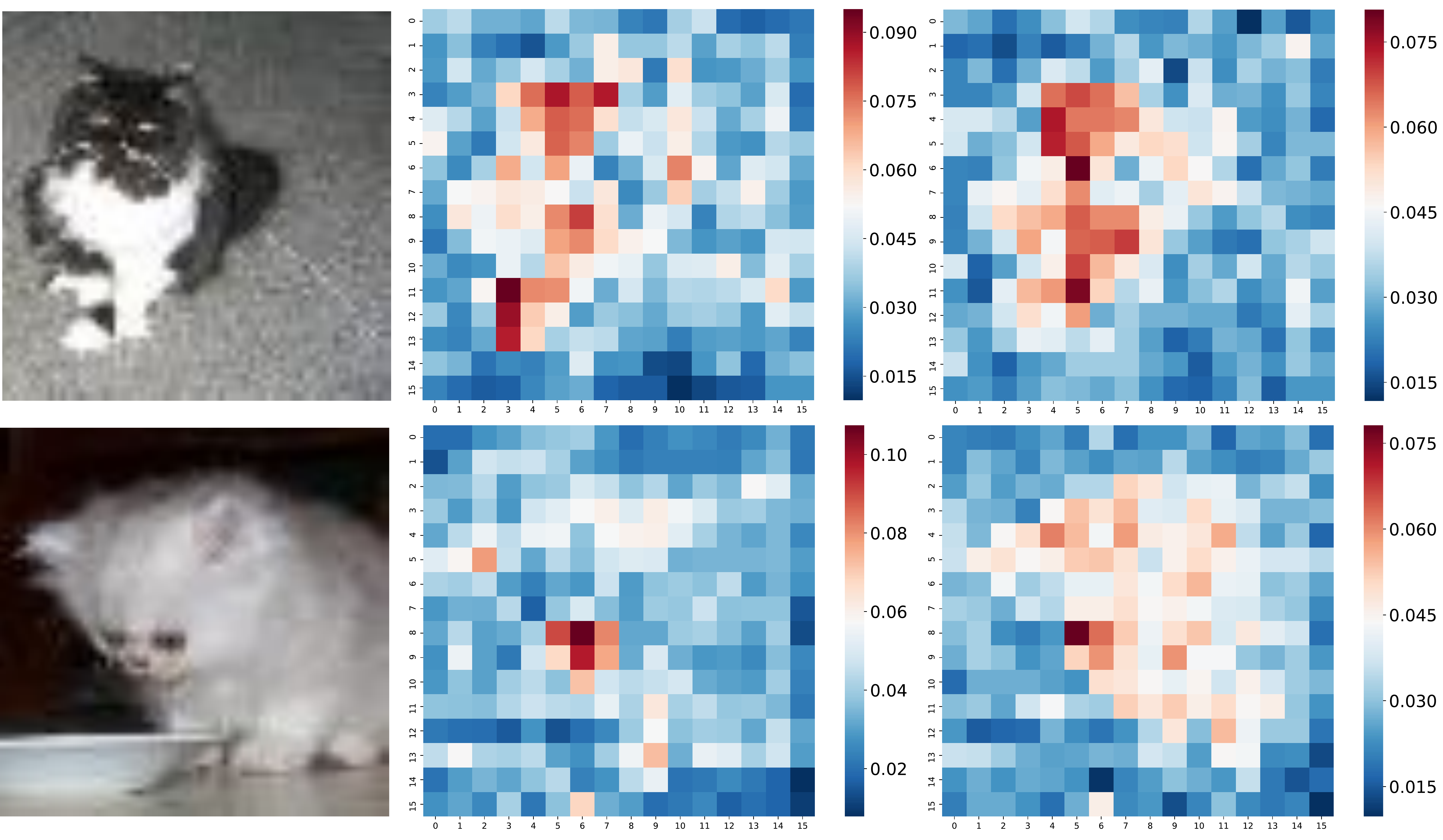} 
  \captionsetup{labelsep=period}
  \vspace{-0.1cm}
  \caption{The Attention maps for SparseSpikformer based on spike firing rate. The token with a redder color contains more significant information while with a bluer color represents less important information.}
\end{figure}

\begin{figure}[t]
\ninept
\begin{minipage}[b]{.496\linewidth} 
  \centering
  %\centerline{\includegraphics[width=4.3cm]{9.pdf}}
  \centerline{\includegraphics[width=1.0\linewidth]{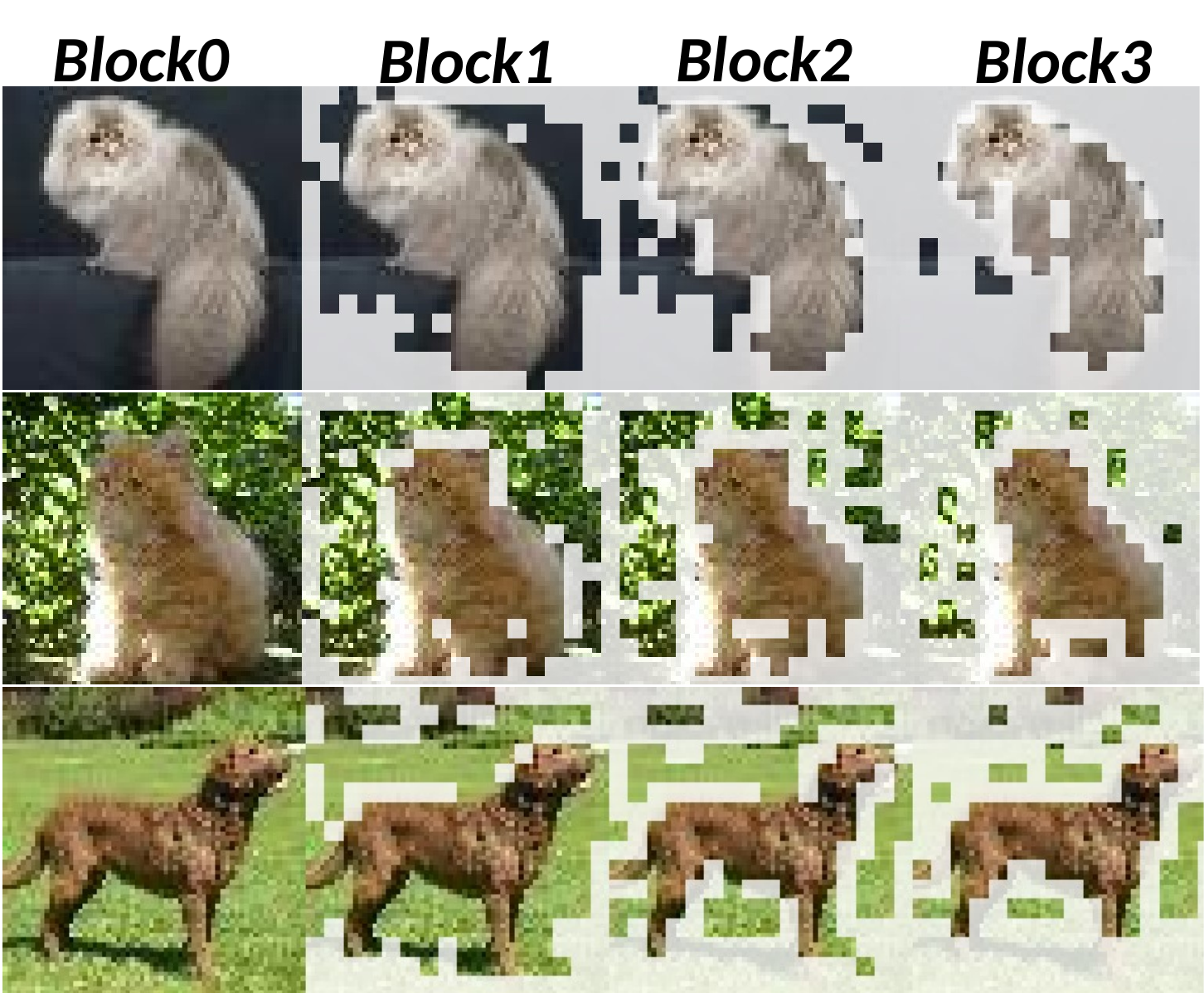}}
\end{minipage}
\hfill
\begin{minipage}[b]{0.496\linewidth}
  \centering
  %\centerline{\includegraphics[width=4.3cm]{10.pdf}}
  \centerline{\includegraphics[width=1.0\linewidth]{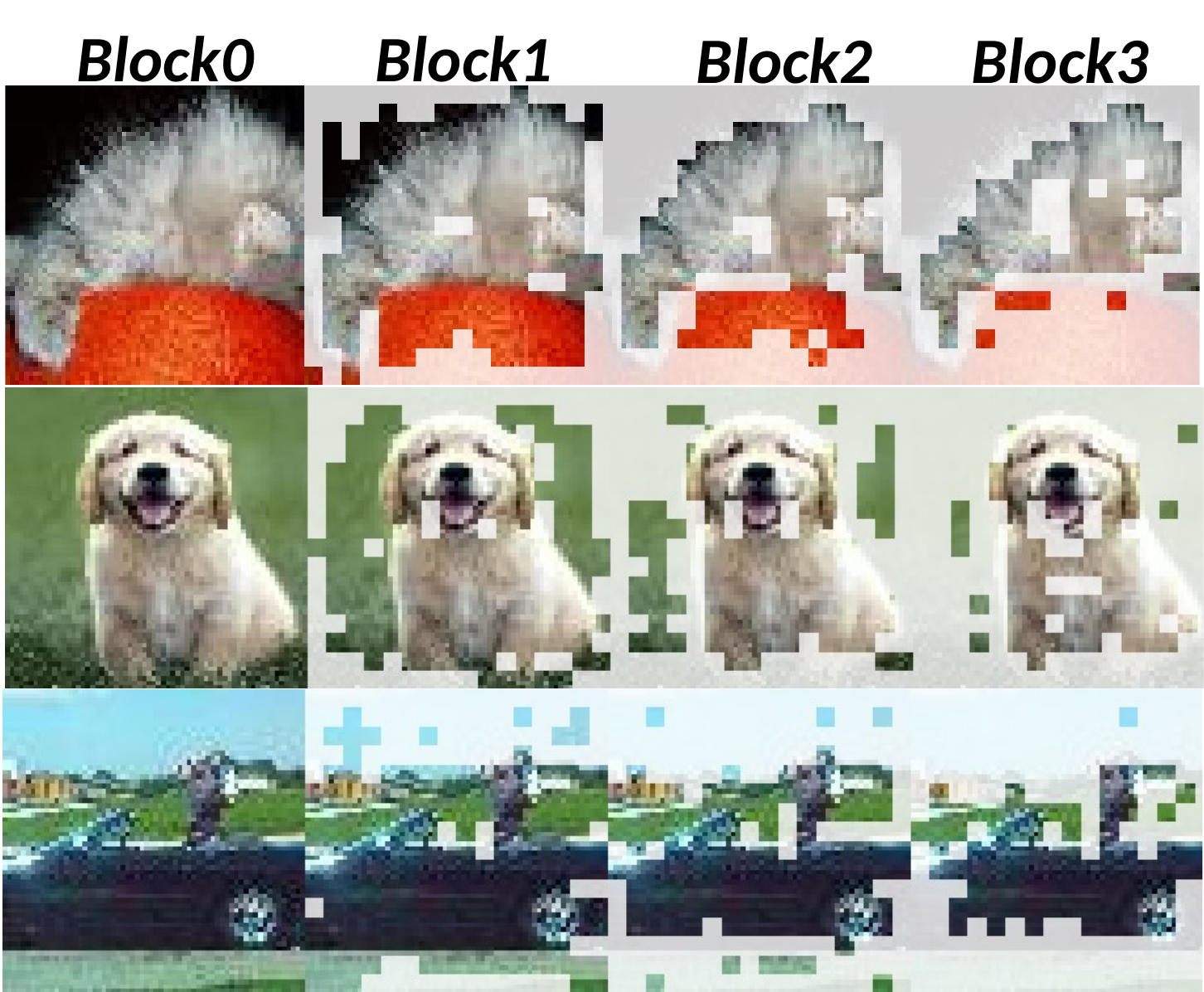}}
\end{minipage}
\captionsetup{labelsep=period}
\caption{Visualization of the token pruning process.}
\end{figure}

% \vspace{-1cm}
\subsection{Dynamic Token Pruning}

In this section, we introduce a lightweight approach to evaluate the importance of image tokens. As shown in Figure 2, we visualize the spike firing rate of the Spiking Self-Attention module, which highlights the importance of different patches within the input image. Specifically, the input image contains some uninformative background tokens, which make a limited contribution to the overall understanding of global information while increasing the computational burden. To overcome this, we incorporate the Spiking Token Select module before the SSA module and gradually drop those useless image tokens according to the predefined pruning level. The detailed processing procedure is demonstrated as follows:

Firstly, we apply global average pooling (GAP) to the input spike sequence $X \in \mathbb{R}^{T \times N \times C}$ 
along the temporal dimension $T$, resulting in $X' \in \mathbb{R}^{N \times C}$ as the input feature.
\begin{equation}
X' = \mathrm{GAP}(X), \quad X' \in \mathbb{R}^{N \times D}, X \in \mathbb{R}^{T \times N \times D}
\end{equation}

Afterwards, the input feature $X'$ is fed into an MLP layer consisting of only two linear layers to generate a series of token probability score maps.
\begin{equation}
S = \mathrm{Softmax}(\mathrm{MLP}(X')), \quad S \in \mathbb{R}^{N \times 2}
\end{equation}
The two-dimensional of $S$ represents the probability scores for keeping or dropping N image tokens, respectively.

In order to make the probability scores differentiable, we use the Gumbel-Softmax \cite{jang2016categorical} function to derive the final keep decisions.
\begin{equation}
\mathbf{D} = \operatorname{Gumbel-Softmax}({S_{*, 1}}), \quad \mathbf{D} \in \{0,1\}^{N} \label{eq:your_label}
\end{equation}

In addition, since the token pruning ratio varies across encoder blocks, it becomes necessary to employ different decisions in different layers. Therefore, as the input data passes through different encoder layers, we need to adjust the current token decision $\hat{D}$ using the Hadamard product $\odot$.
\begin{equation}
\hat{\mathbf{D}} = \hat{\mathbf{D}} \odot {\mathbf{D}} \label{eq:your_label}
\end{equation}

\vspace{-\baselineskip}
\begin{table*}[htb]
\ninept
\centering
\captionsetup{labelsep=period}
\caption{Performance of models under different keep ratios.}
\vspace{-0.2cm}
\ninept
\setlength{\tabcolsep}{2.1pt}
\begin{tabular}{cccccccccccccc}
%\hline
\specialrule{.1em}{.1em}{.1em}
\multirow{2}{*}{\textbf{Model}} & \multirow{2}{*}{\textbf{\begin{tabular}[c]{@{}c@{}}Keep\\ ratio\end{tabular}}} & \multicolumn{3}{c}{\textbf{CIFAR10}} & \multicolumn{3}{c}{\textbf{CIFAR100}} & \multicolumn{3}{c}{\textbf{CIFAR10-DVS}} & \multicolumn{3}{c}{\textbf{DVS-128}} \\ \cmidrule(lr){3-5}\cmidrule(lr){6-8}\cmidrule(lr){9-11}\cmidrule(lr){12-14}
& & Acc & Throughput & FLOPs & Acc & Throughput & FLOPs & Acc & Throughput & FLOPs & Acc & Throughput & FLOPs \\
\hline
Spikformer \cite{zhou2022spikformer} & 1.0 & 94.95 & 950.28 & 3.74G & 77.28 & 1049.55 & 3.74G & 78.3 & 322.18 & 7.78G & 98.26 & 304.25 & 7.78G \\
\hline
\multirow{5}{*}{\begin{tabular}[c]{@{}c@{}}Sparse\\ Spikformer\end{tabular}} & 0.9 & 95.22 & 1293.32 & 3.44G & 77.81 & 1156.07 & 3.44G & 79.1 & 336.84 & 7.36G & 98.26 & 315.52 & 7.36G \\
& 0.8 & 95.18 & 1381.73 & 3.24G & 77.70 & 1245.03 & 3.24G & 79.3 & 340.46 & 6.93G & 97.91 & 331.87 & 6.93G \\
& 0.7 & 95.01 & 1448.71 & 3.04G & 77.42 & 1324.60 & 3.04G & 79.3 & 347.64 & 6.57G & 97.91 & 346.73 & 6.57G \\
& 0.6 & 95.03 & 1521.58 & 2.88G & 77.09 & 1394.16 & 2.88G & 78.4 & 374.96 & 6.27G & 97.91 & 360.18 & 6.27G \\
& 0.5 & 94.77 & 1586.64 & 2.75G & 76.78 & 1455.03 & 2.75G & 79.1 & 408.90 & 6.03G & 97.91 & 371.13 & 6.03G \\ 
\specialrule{0.1em}{0em}{0em}
\end{tabular}
\end{table*}

\vspace{-0.1cm}
\subsection{Weight Parameters Pruning}

Although Spikformer has demonstrated remarkable performance on various datasets, its sparse nature of SNN implies that only a fraction of neurons fire spikes at any given moment, resulting in potential wastage of computing resources. To address this issue, we hypothesize that non-spiking neuron connections can be removed via pruning techniques as a means of reducing storage space resources.

As an efficient neural network pruning method, LTH can discover the winning tickets in both ANN and SNN models, making it much sparse \cite{kim2022exploring}. Besides, the accuracy of the subnetwork can be further improved through Iterative Magnitude Pruning (IMP) \cite{frankle2019stabilizing} and retraining. In the first iteration of the pruning process, LTH generates a dense neural network model $f(x;\theta)$ using conventional random initialization and trains it until convergence. Then, the weight parameters are sorted, and the $p\%$ of the connections with the smallest absolute values are removed. After that, the remaining weight parameters are reinitialized, resulting in the formation of the subnetwork $f(x;M\odot\theta)$, where the mask $M \in \{0, 1\}$ denotes the pruning or preservation of the connectivity relationships from the original network $f(x;\theta)$. The pruned subnetwork is subsequently retrained in the next iteration, and this iterative process will be repeated for $K$ times. Ultimately, the resulting subnetwork prunes $p^K\%$ of the weights, leading to a more sparse model. In our experiments, we set the values of $p$ and $K$ to 25 and 15.

\vspace{-0.4cm}
\section{Experiments}
\label{sec:experiments}

\vspace{-0.2cm}
\subsection{Experimental Settings}

We evaluate SparseSpikformer on four publicly available datasets, including static datasets (CIFAR-10, CIFAR-100) \cite{2009Learning} and neuromorphic datasets (CIFAR10-DVS \cite{li2017cifar10}, DVS-128 \cite{amir2017low}). During the training process, we insert the token selector after the 2$^{nd}$, 3$^{rd}$, and 4$^{th}$ layers of the encoder block. Moreover, we also conduct a series of experimental comparisons with different sparsities. We follow the training principles and optimization methods proposed in the original Spikformer \cite{zhou2022spikformer}. All the experiments are conducted on the NVIDIA Tesla V100 with a single GPU.

\begin{table*}[htb]
\ninept
\centering
\captionsetup{labelsep=period}
\caption{Accuracy comparison across different weight sparsity levels.}
\vspace{-0.2cm}
\ninept
\begin{tabular}{ccccccccc}
\specialrule{.1em}{.1em}{.1em}
% \hline
\multirow{2}{*}{\textbf{Model}}                                                    & \multicolumn{2}{c}{\textbf{CIFAR10}} & \multicolumn{2}{c}{\textbf{CIFAR100}} & \multicolumn{2}{c}{\textbf{CIFAR10-DVS}} & \multicolumn{2}{c}{\textbf{DVS-128}} \\ \cline{2-9}
                                                                                   & Sparsity          & Top-1 Acc         & Sparsity          & Top-1 Acc          & Sparsity            & Top-1 Acc           & Sparsity             & Top-1 Acc             \\ \hline
\multirow{3}{*}{Spikformer \cite{zhou2022spikformer}}                                                        & 75.90\%           & 94.78            & 75.90\%           & 77.71             & 81.3\%              & 77.1               & 85.7\%               & 98.26                \\
                                                                                   & 86.22\%           & 94.91            & 86.22\%           & 76.53             & 89.0\%              & 76.8               & 93.4\%               & 98.26                \\
                                                                                   & 93.90\%           & 94.34            & 92.03\%           & 75.29             & 95.8\%              & 75.9               & 95.8\%               & 98.26                \\ \hline
\multirow{3}{*}{\begin{tabular}[c]{@{}c@{}}SparseSpikformer \\ $\rho$=0.7\end{tabular}} & 73.26\%           & 95.28            & 75.90\%           & 77.48             & 81.3\%              & 77.9               & 85.7\%               & 98.95                \\
                                                                                   & 88.74\%           & 95.06            & 86.23\%           & 76.81             & 89.0\%              & 78.5               & 93.4\%               & 98.26                \\
                                                                                   & 93.90\%            & 94.97            & 92.04\%           & 75.12             & 95.8\%              & 76.2               & 95.8\%               & 98.26                \\ 
\specialrule{0.1em}{0em}{0em}
\end{tabular}
\end{table*}

\vspace{-0.3cm}
\subsection{Token Pruning Results}

In Table 1, we report the results achieved by SparseSpikformer on different keep ratios $\rho$ and multiple datasets. These results encompass three key metrics, including Top-1 accuracy, Throughput, and FLOPs. It is obvious that SparseSpikformer demonstrates a significant reduction in floating-point calculation costs by 22.49\% to 26.47\%, as well as an accelerated inference runtime by 21.98\% to 66.96\% while preserving minimal loss in accuracy. The experimental results unmistakably highlight the exceptional performance of SparseSpikformer, which provides support for deploying applications in resource-constrained environments.

% Specifically, during the experiments, we gradually drop some tokens before each Encoder block, preserving different proportions of image tokens such as [$\rho$, $\rho^{2}$, $\rho^{3}$]. In Figure 3, we visualize the pruning process, where we set $\rho$=0.7.

Throughout the experiments, we gradually remove a fraction of tokens before each block, preserving distinct ratios of image tokens, such as [$\rho$, $\rho^{2}$, $\rho^{3}$]. Figure 3 provides a visual representation of this pruning process, with $\rho$=0.7.

\vspace{-0.3cm}
%\vspace{-\baselineskip}
\subsection{Weight Pruning Results}

% Table 2 presents the accuracy comparison of SparseSpikformer and Spikformer under various weight sparsity levels. The sparsity refers to the percentage of weights pruned from the overall parameters. In this experiment, we maintain a fixed token keep rate $\rho$ of 0.7 for SparseSpikformer. The results indicate that the LTH can successfully identify winning tickets even when 90\% of the weights are pruned in both Spikformer and SparseSpikformer models, leading to outstanding performance. It is particularly noteworthy that SparseSpikformer can achieve remarkably high weight sparsity levels ($p$=95.8\%) on neuromorphic datasets such as CIFAR10-DVS and DVS-128. Furthermore, based on the data presented in Tables 1 and 2, it is obvious that SparseSpikformer (Table 2) achieves a remarkable balance between accuracy and efficiency, closely matching the accuracy of the original Spikformer (Table 1) even with a 90\% reduction in weight parameters. This reduction significantly alleviates the challenges associated with weight storage.
\vspace{-0.1cm}
In Table 2, we compare the accuracy of SparseSpikformer and Spikformer models at different levels of weight sparsity. The "sparsity" refers to the percentage of weights that have been pruned from the overall parameters. For our experiment, we maintain a fixed token keep rate $\rho$ of 0.7 for SparseSpikformer. Our results show that LTH can successfully identify winning tickets in both the Spikformer and SparseSpikformer models, leading to outstanding performance. Additionally, based on the data presented in Tables 1 and 2, it is clear that SparseSpikformer (Table 2) achieves a remarkable balance between accuracy and efficiency, closely matching the accuracy of the original Spikformer (Table 1) even with a 90\% reduction in weight parameters. This reduction significantly alleviates the challenges associated with weight storage.

\begin{table}[tb!]
\setlength{\tabcolsep}{1.8pt}
\centering
\ninept
%\footnotesize
\captionsetup{labelsep=period}
\caption{Comparisons with different pruning methods.}
\vspace{-0.2cm}
\ninept
\begin{tabular}{cccccc}
\specialrule{.1em}{.1em}{.1em}
%\hline
\multirow{2}{*}{\textbf{Method}}                                                      & \multicolumn{1}{l}{\multirow{2}{*}{\textbf{\begin{tabular}[c]{@{}l@{}}$\rho$\end{tabular}}}} & \textbf{CIFAR10} & \textbf{CIFAR100} & \textbf{CIFAR10-DVS} & \textbf{DVS-128} \\ \cline{3-6}                                                                                       & \multicolumn{1}{l}{}                                                                                   & Top-1 Acc         & Top-1 Acc          & Top-1 Acc             & Top-1 Acc             \\ \hline
\multirow{3}{*}{\begin{tabular}[c]{@{}c@{}}Random \\ Selector\end{tabular}}           & 0.8                                                                                                    & 94.20            & 73.11             & 78.9                 & 97.56                \\
                                                                                      & 0.7                                                                                                    & 94.07            & 72.72             & 78.9                 & 97.56                \\
                                                                                      & 0.6                                                                                                    & 93.02            & 70.32             & 78.2                 & 97.56                \\ \hline
\multirow{3}{*}{\begin{tabular}[c]{@{}c@{}}Spiking \\ Token \\ Selector\end{tabular}} & 0.8                                                                                                    & 95.18            & 77.70             & 79.3                 & 97.91                \\
                                                                                      & 0.7                                                                                                    & 95.01            & 77.42             & 79.3                 & 97.91                \\
                                                                                      & 0.6                                                                                                    & 95.03            & 77.07             & 78.4                 & 97.91                \\ 
\specialrule{0.1em}{0em}{0em}
\end{tabular}
\end{table}

\vspace{-0.2cm}
\subsection{Ablation Analysis }
\vspace{-0.1cm}
\textbf{Effectiveness of token pruning techniques}. To verify the effectiveness of the spiking token selector module, we compare two different pruning strategies across multiple datasets. The results are summarized in Table 3. Specifically, we employ the random selector to randomly prune image tokens, while the spiking token selector is adopted to selectively prune tokens with the lowest spike firing rates. The experimental results clearly indicate that the spiking token selector module can achieve a better performance than the random selector. Based on that, we can infer that our approach is effective for dynamic token pruning.

\noindent\textbf{Effectiveness of weight pruning techniques}. In Figure 4, we present the results of our experiments to evaluate the performance of different LTH techniques, including IMP and EB \cite{you2019drawing}, on the CIFAR10 and CIFAR100 datasets. The results clearly demonstrate that both IMP and EB techniques are able to identify the winning tickets within the original network. However, when evaluating the performance of subnetworks with high weight sparsity (e.g. 75.90\%, 86.22\%, and 92.04\%), IMP consistently outperforms EB in terms of Top-1 accuracy, achieving superior results by 3.52\% to 9.74\%. 

Furthermore, during the training process, we also compare the effects of network re-initialization techniques using Random re-initialization (RR) and Rewinding (IMP). The experiments indicate that the accuracy of the winning ticket through Rewinding is closer to that of the original network.

\begin{figure}[t]
\ninept
\begin{minipage}[b]{.48\linewidth} 
  \centering
  \centerline{\includegraphics[width=4.9cm]{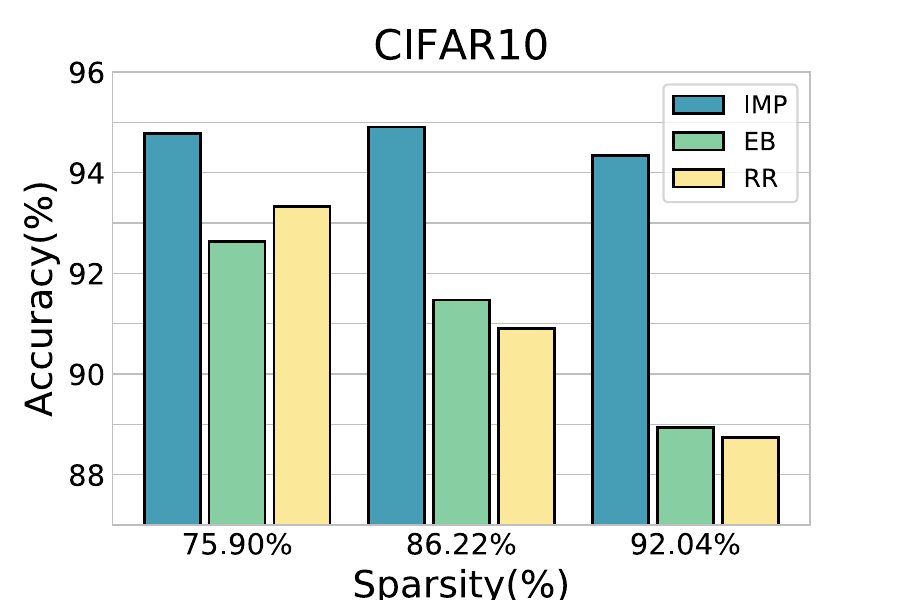}}
  \centerline{(a) CIFAR10}\medskip
\end{minipage}
\hfill
\begin{minipage}[b]{0.48\linewidth}
  \centering
  \centerline{\includegraphics[width=4.9cm]{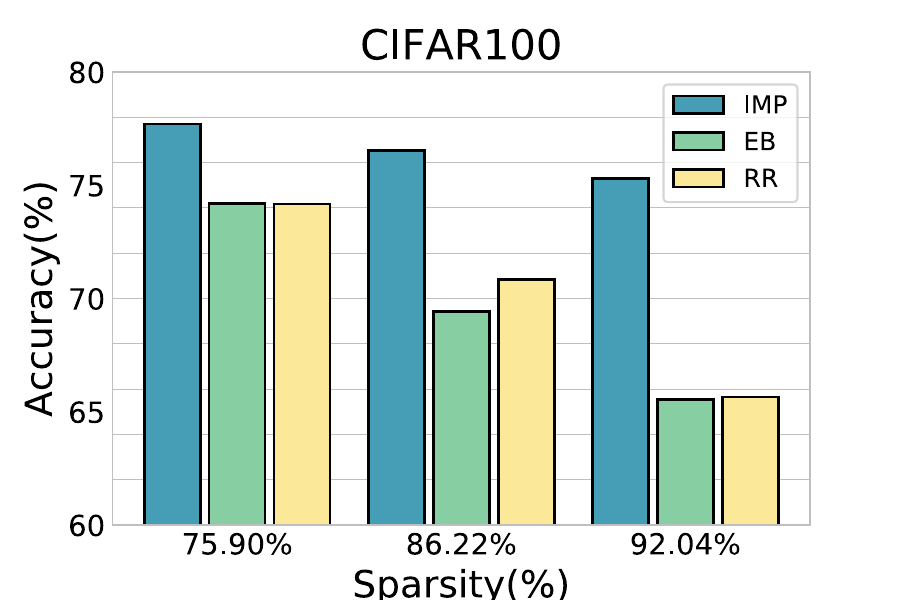}}
  \centerline{(b) CIFAR100}\medskip
\end{minipage}
\captionsetup{labelsep=period}
\vspace{-0.3cm}
\caption{Illustration of the accuracy of Iterative Magnitude Pruning (IMP), Early-Bird (EB), and the Random Re-initialization (RR) techniques in LTH.}
\end{figure}

\vspace{-0.2cm}
\section{Conclusion}
\label{ssec:subhead}

\vspace{-0.1cm}
In this paper, we introduce SparseSpikformer, a sparse and hardware-friendly model based on Spikformer architecture. By incorporating the Spiking Token Selector module and the LTH weight pruning technique, we can effectively achieve a balance between accuracy and hardware resource trade-offs. The experimental results demonstrate that our model performs well on both static and neuromorphic datasets. Additionally, SparseSpikformer also accelerates runtime during the inference phase and cuts down the floating-point calculations, which provides possibilities for implementation on edge devices.

\vfill\pagebreak

\ninept
\bibliographystyle{IEEEbib}
\bibliography{strings,refs}

\end{document}